\documentclass[letterpaper, 10 pt, conference]{ieeeconf}  % Comment this line out if you need a4paper

\IEEEoverridecommandlockouts                              % This command is only needed if 
\usepackage{balance} 
\usepackage{amssymb}
\usepackage{graphics}
\usepackage{graphicx}
\usepackage{amsmath} % assumes amsmath package installed
\usepackage{floatrow}
\usepackage{url}
\usepackage{svg}
\usepackage{textcomp}
\usepackage{gensymb}
\usepackage[ruled,vlined]{algorithm2e}
\usepackage{multirow}
\usepackage{tablefootnote}

\usepackage{multicol}

\title{\LARGE \bf
Perception-Intention-Action Cycle in Human-Robot Collaborative Tasks
}

\author{J. E. Dom\'{i}nguez-Vidal, Nicol\'{a}s Rodr\'{i}guez, Ren\'{e} Alqu\'{e}zar and Alberto Sanfeliu% <-this % stops a space
\thanks{Work supported under the European project CANOPIES (H2020- ICT-2020-2-101016906) and the MINECO/AEI ROCOTRANSP project (PID2019-106702RB-C21 / AEI /10.13039/501100011033). The first author acknowledges Spanish FPU grant with ref. FPU19/06582.}% <-this % stops a space
\thanks{The authors are with the Institut de Rob\`otica i Inform\`atica Industrial (CSIC-UPC). Llorens Artigas 4-6, 08028 Barcelona, Spain. {\tt\footnotesize  \{jdominguez, nrodriguez, ralquezar, sanfeliu\}@iri.upc.edu}}%
}

\begin{document}

\maketitle
\thispagestyle{empty}
\pagestyle{empty}

%%%%%%%%%%%%%%%%%%%%%%%%%%%%%%%%%%%%%%%%%%%%%%%%%%%%%%%%%%%%%%%%%%%%%%%%%%%%%%%%
\begin{abstract}

In this work we argue that in Human-Robot Collaboration (HRC) tasks, the Perception-Action cycle in HRC tasks can not fully explain the collaborative behaviour of the human and robot and it has to be extended to Perception-Intention-Action cycle, where Intention is a key topic. In some cases, agent Intention can be perceived or inferred by the other agent, but in others, it has to be explicitly informed to the other agent to succeed the goal of the HRC task. The Perception-Intention-Action cycle includes three basic functional procedures: Perception-Intention, Situation Awareness and Action. The Perception and the Intention are the input of the Situation Awareness, which evaluates the current situation and projects it, into the future situation. The agents receive this information, plans and agree with the actions to be executed and modify their action roles while perform the HRC task. In this work, we validate the Perception-Intention-Action cycle in a joint object transportation task, modeling the Perception-Intention-Action cycle through a force model which uses real life and social forces. The perceived world is projected into a force world and the human intention (perceived or informed) is also modelled as a force that acts in the HRC task. Finally, we show that the action roles (master-slave, collaborative, neutral or adversary) are intrinsic to any HRC task and they appear in the different steps of a collaborative sequence of actions performed during the task.

\end{abstract}

%%%%%%%%%%%%%%%%%%%%%%%%%%%%%%%%%%%%%%%%%%%%%%%%%%%%%%%%%%%%%%%%%%%%%%%%%%%%%%%%
\section{Introduction}

Collaboration is a process where two or more agents work together as partners to achieve a shared goal \cite{terveen1995overview}. Applied to robotics, Human-Robot Collaboration (HRC) is a special type of Humna-Robot Interaction (HRI) where at least one robot and one human perceive, predict, share their intentions, and then act accordingly. Therefore, collaboration is a key challenge to enhance human capabilities in a robotic society. 

The Perception-Action cycle has served as a framework for the development and understanding of artificial intelligence systems as well as robotics. Early works in robotics, assume that a traditional decomposition of functionalities starts from perception and finalize in a sequence of robotic actions~\cite{brooks1986robust}. This means that the perception and understanding of the environment in which a robot operates is essential for it to be able to navigate, select the right tool or, in general, to perform its task effectively by making the right decisions at the right time~\cite{goldhoorn2017searching}. However, when this task must be performed collaboratively with one or more humans, it is no longer sufficient to perceive and understand the environment. It is necessary to know the human's intention.

\begin{figure}
    \centering
    \includegraphics[width=0.95\textwidth]{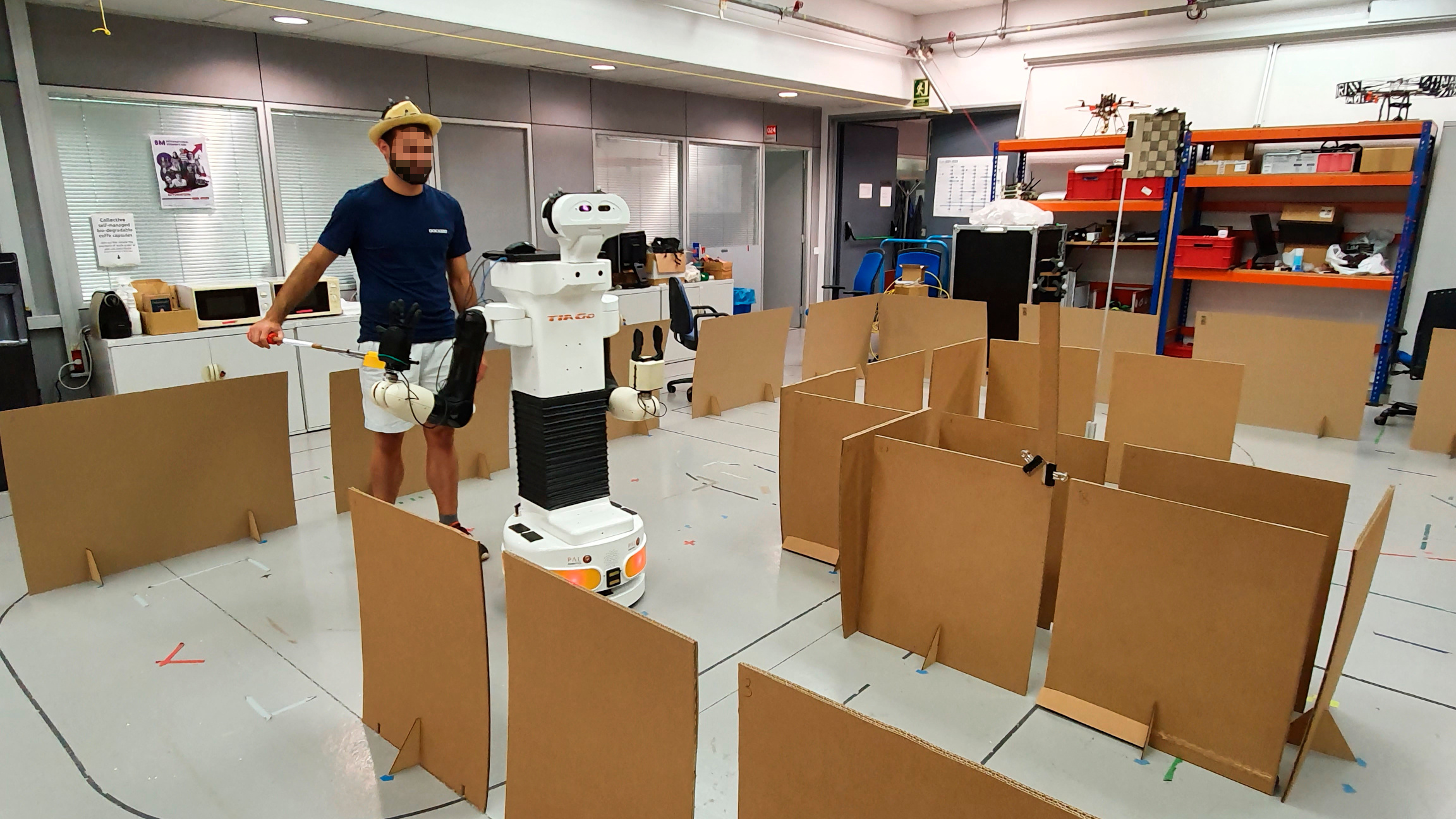}
    \caption{{\bf Example of Human-robot pair collaboratively transporting an object.} Both agents must navigate through a complex environment with multiple walls. Human has explicitly expressed his intention of moving himself behind the robot to pass through a narrow passage. The transported object is an aluminium bar.}
    \label{fig:human-robot_pair_experiment}
\end{figure}

It can be argued that it is possible to interpret the human's intention by perceiving their actions. However, the myriad of misunderstandings which we humans make when we try to interpret the intention of our fellow humans from their actions, demonstrates the need to directly elicit this intention for the correct development of multiple tasks and consider it as another element of the decision-making cycle. Especially if the agents have different representations of the world which may hinder the interpretation process, as occurs in a human-robot pair due to the multifaceted ways~\cite{johnson1989mental} a human can model the perceived information.

With this in mind, and focused in the HRC paradigm, we propose as a first contribution a review of the Perception-Action cycle by incorporating the human's intention at the same level of perception and using the concept of Situation Awareness~\cite{Endsley2000} to combine and process all the information. This allows for a better decision making selecting or adapting first the agents their roles and then making their plans. As a second contribution, we propose to combine physical and virtual forces for modeling HRI tasks. More specifically, we integrate the physical force exerted by the human, their intention modeled as a virtual force and the environment modeled as a combination of virtual forces using a simple yet effective force-based model as the Social Force Model~\cite{Helbing1995}. Finally, we perform a series of human-robot collaborative object transportation experiments to validate the proposal.

In the reminder of the paper, we start describing the relevant related works in Section~\ref{sec:related_work}. In Section~\ref{sec:PIA_cycle} we present the Perception-Intention-Action Cycle as an extended framework to tackle collaborative tasks. Later, Section~\ref{sec:PIA_model} presents the model used in the experiments to validate the framework and Section~\ref{sec:roles_and_strategies} comments the implications of using this model. Finally, Section~\ref{sec:experiments} and~\ref{sec:conclusions} present the conducted experiments and the conclusions, respectively.

\section{Related work}\label{sec:related_work}

Early works in robotics use the Perception-Action cycle to decompose the functional modules of the robot control~\cite{brooks1986robust},~\cite{albus75},~\cite{albus1993}. This allowed the design and development of more complex robots~\cite{Nishiwaki2000} and control architectures based on how the human brain processes~\cite{cutsuridis2013} to improve robotic capabilities to perform specific tasks. However, when it comes to include the human in-the-loop, authors recognize that it is not enough to obtain human-like robots~\cite{lee2018}. That is why we extended it including the human's intention.

Situation Awareness~\cite{Endsley2000},~\cite{endsley2000direct} is according to the author the knowledge of what is going on around you. In other words, to sift all the irrelevant stimuli and understand which information is important to attend. Originally used in aviation, it has long been recognized as a core competence for intelligent behavior and correct decision-making, specially in critical combat environments. It has three levels~\cite{Scholtz2002} going from (1) just perceiving the surrounding information and (2) integrating the different information sources according to their relevance to (3) make future predictions based on the comprehension of the current situation. Despite the power of this concept, to the best of our knowledge, it was only used in robotics to design user interfaces~\cite{riley2004},~\cite{riley2010},~\cite{opiyo2021} but not as a core component in the robot's reasoning to understand the intention of its human partner as in our work.

Speaking of intention,~\cite{Losey2018} makes a review of measuring intent in physical HRI and its interpretation by the robot to establish a shared-control policy, which is named as role allocation. The concept of role is known in the literature going from the classical master-slave and collaborative options~\cite{Mortl2012} to the less known adversarial or antagonistic case~\cite{Jarrasse2012}. Examples of the use of roles to model or condition the robot's behaviour are the aforementioned~\cite{Mortl2012} in collaborative transportation or~\cite{Dalmasso_ICRA2020} in collaborative search. This last work is similar to our in the sense that they also obtain the human's intention and use it to condition the robot's action. Similarly, they also use a force-based model (based on~\cite{Helbing1995}) to represent the scenario in which they perform the task.

\section{Perception-Intention-Action Cycle}\label{sec:PIA_cycle}

\begin{figure*}[ht]
    \centering
    \includegraphics[width=0.90\textwidth]{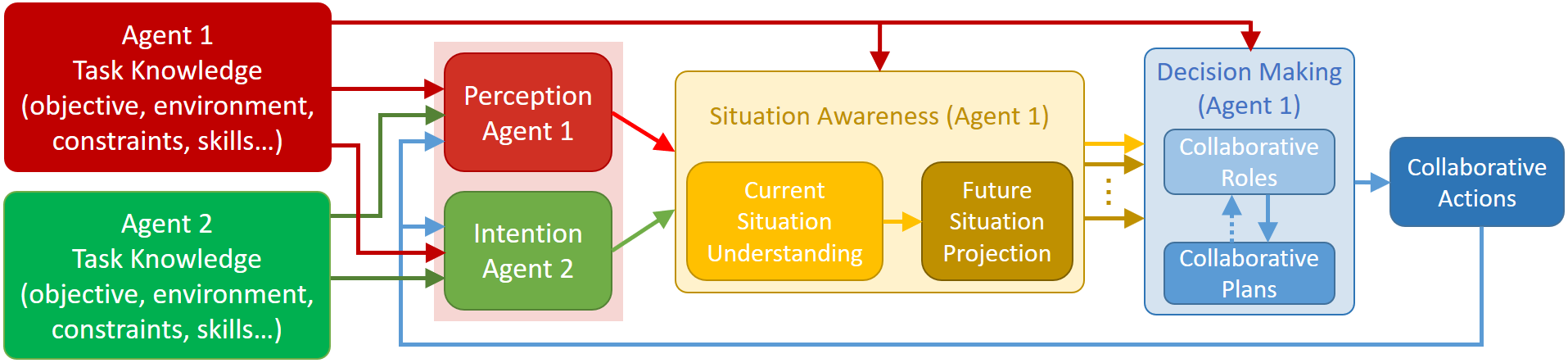}
    \caption{{\bf General information flow in each agent in a collaborative task.} Available information about the task is obtained as well as the intention that the same information from the point of view of both agents generates in the other agent. The situation awareness comprehends the current situation and projects into the future situation. This projection allows each agent to establish collaborative plan according to the role each agent is showing at the moment. This plan generates the following actions which are perceived again initiating a new cycle.}
    \label{fig:block_diagram}
\end{figure*}

When a robot is navigating in an urban environment surrounded by humans, it can interpret each human as a moving obstacle, estimate their velocity and acceleration, and with this information make an estimate of the human's movement, typically with increasing uncertainty over time. However, if the robot knew the human's intention, i.e. where they want to go, the above calculation would be greatly simplified and the uncertainty would be considerably lower.

As mentioned in the introduction, this intention may not always be perceived. Imagine the reader two humans collaboratively carrying an object, for example, a table. If one of them starts to turn, the other does not know whether they are doing so because they actually want to change direction or because they are turning sideways in order to pass through a narrow passage. If the object being transported is also heavy, when one of them wants to stop, they will not exert an opposing force to the other because inertia would make this movement dangerous. What they typically do is to continue with the inertia and talk to their partner to indicate their intention to stop. In both cases it is necessary to explicitly state what they want to do to eliminate uncertainties and allow the task to proceed correctly. Figure~\ref{fig:human-robot_pair_experiment} shows a situation in which the human must explicitly tell the robot whether he wants to change direction or to move behind it to surpass an obstacle.

The adversarial case is also of special interest despite being typically ignored in robotics due to its almost infinite casuistry. This occurs when one of the agents not only does not collaborate with the task but the goal of their task is contrary to that of the other agent's task. Let's take for example a professional tennis match. Trying to estimate the opponent's next shot based on their positioning may not be enough as they may be resorting to deception. Having studied their playstyle, on the other hand, allows us to know their real intention, which can make the difference between winning and losing.

All of the above (including the possibility of being able to consider the human as an adversary if they are behaving as such) motivates us to extend the classical Perception-Action cycle by including the human's intention according to the framework shown in Fig.~\ref{fig:block_diagram}.

The initial assumptions are that there are a minimum of two agents and that there is a collaborative task in which both agents need to participate. Each agent possesses their own knowledge of the task to be performed including the goal of their task or the constraints, both those of the task itself (time, maximum number of attempts...) and of the agent themselves (their height, the number of available limbs/actuators, skills...). In turn, there is also a knowledge about the scenario in which the task takes place which may already be known by each agent or perceived through their sensors (sight, hearing, RGB camera, LiDAR...). This same perception is also responsible for making each agent to detect the changes occurring in the environment, the constraints or even the goal of the task. However, each agent can receive partial and, therefore, different information as well as represent this information differently. This is why the intention of the other agent must be taken into account when making any decision, since each agent does not usually have access to the representation of the information that the other agent is making. Note that this intention can be expressed implicitly (through the actions performed by the other agent and, therefore, inferable using the own knowledge) or explicitly independently of the action which moves the task forward (saying out loud to your partner that you want to get behind them to pass through a narrow passage).

With one's own perception of the world and the intention of the other agent, situation awareness comes into play. This concept, presented by Endsley and Garland in~\cite{Endsley2000}, is originated in the field of aviation and is used to explain the mental process of a pilot in a combat situation. In general, it implies knowing and understanding what is going on around oneself. With this concept we can, from the information received and using the previous own knowledge, understand the current situation and make a projection of the future one. This projection should be understood not as a single prediction but as a probability distribution of the possible future situations. 

This projection is used in a decision making process firstly to know the role which each agent intends to exercise based on their intention. For example, if the other agent intends to follow the plan proposed by the first one, they will be assigned a slave role while, if their intention goes against the development of the task, they will be assigned an adversarial role. More information about the roles we consider will be presented in Section~\ref{sec:roles_and_strategies}. Once the role assigned to each agent is known, a joint plan to be executed by both agents can be planned. This process can be executed several times if we are analyzing every possible prediction trying to find an action to make the other agent to act on a different way or just once if we are trying to adapt ourselves to the most probable future situation. Finally, this collaborative plan is converted into specific actions to be executed by each agent which result is perceived to initiate a new cycle.

Applied to robotics, both the situation awareness block and the role allocation can be performed with a rudimentary state machine, a classical Markov decision process or the most recent architectures based on artificial neural networks. In this way, this framework allows us to extend the classic Perception-Action cycle to unify it with Theory-of-Mind concepts as well as works based on understanding the roles which arise between a human and a robot when performing collaborative tasks such as that of Mörtl et al.~\cite{Mortl2012} through the concept of situation awareness.

\section{Perception-Intention-Action Force-based Model}\label{sec:PIA_model}

\begin{figure*}[ht]
    \centering
    \includegraphics[width=0.98\textwidth]{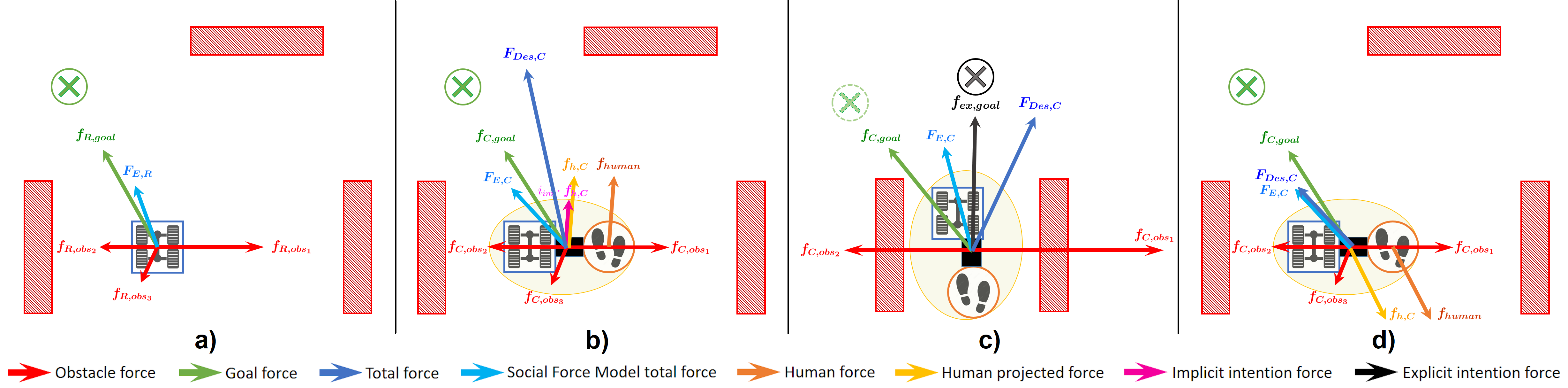}
    \caption{{\bf Perception-Intention-Action Force-based Model applied in different situations.} A - Robot navigating alone through obstacles. B - Human-robot pair navigating collaboratively and human making force to fulfill the task. C - Human-robot pair passing collaboratively trough a narrow pass. D - Human-robot pair navigating collaboratively and human making force against what would be expected to fulfill the task. Forces not normalized for clarity.}
    \label{fig:force_based_model}
\end{figure*}

As mentioned in the introduction, we will use a collaborative rigid object transportation task as a use case to demonstrate our proposal. For simplicity, we will consider the number of implied agents $N=2$ being the robot the agent $1$ and the human the agent $2$. Our objective is to create a model which allows us to compare and combine the real force exerted by the human with virtual forces generated to understand the environment. We will use a global planner to calculate the robot's original plan as a succession of waypoints or partial goals to reach the task's goal. If they were two robots collaborating, a shared planner would be enough but, since there is a human in the loop, we will try to understand their actions to condition and select the final actions to be performed by the robot. For that, we will use a force sensor attached to the robot's wrist which is in contact with the transported object. 

In this way, we can take~\cite{Mortl2012} as a starting point. In that work, a rigid object is also transported collaboratively among several agents, so we will attach a frame $C$ to an arbitrary point over the object and assume that the dynamics of the object in that frame are determined by the joint action of two forces: the force exerted by the robot at one end of the object and the force exerted by the human at the other end. In turn, the force exerted by the robot is determined by its interpretation of the environment (present obstacles, location of the task's goal...). Therefore, we can define the desired force to be exerted on frame C as:

\begin{equation}
	\label{eq:1}
	\boldsymbol{F_{Des, C}} = \boldsymbol{F_{E, C}} + \boldsymbol{F_{H, C}}
\end{equation}

\noindent
being $\boldsymbol{F_{E, C}}$ the component due to the environment and $\boldsymbol{F_{H, C}}$ the component due to the force exerted by the human.

\subsection{Environment perception}

In order to model the task's environment using virtual forces, we can start from~\cite{Helbing1995}. According to this work, the accelerations and decelerations of a passer-by walking along a busy street are determined by the joint action of virtual repulsive and attractive forces according to the following expression:

% We can model the task's environment using virtual forces like in~\cite{Helbing1995}. According to this work, the accelerations and decelerations of a passer-by walking along a busy street are determined by the joint action of virtual repulsive and attractive forces according to the following expression:

\begin{equation}
	\label{eq:2}
	\boldsymbol{f_{a,o}} = U_{a,o} \left ( \left \| \boldsymbol{r_{a,o}} \right \| \right )
\end{equation}

\noindent
being the virtual repulsive (attractive) force generated by object $o\in O$ over the agent $a\in \left \{r, h  \right \}$, $\boldsymbol{f_{a,o}}$, the result of applying a monotonic decreasing (increasing) potential $U_{a,o}$ over the vector $\boldsymbol{r_{a,o}}$ which goes from the agent to the object.

These objects $o$ can be obstacles or the goal. In the first case, we can consider that all the obstacles in the robot's field of view will generate a virtual repulsive force decreasing with the distance from a maximum $f_{rep, max}$. If the obstacle is further than a threshold $d_{max}$ we will not consider it:

% In this way, we can consider that all the obstacles in the robot's field of view will generate a virtual repulsive force decreasing with the distance from a maximum $f_{rep, max}$  to $0$ at a maximum distance $d_{max}$:

\begin{equation}
    \label{eq:3}
    \boldsymbol{f_{a,obs}} =
        \begin{cases}
        f_{rep, max} \cdot e^{ -\frac{\left \| \boldsymbol{r_{a,obs}} \right \|}{d_{max}} } & \text{if $\boldsymbol{\left \| r_{a,obs} \right \| } < d_{max}$}\\
        0 & \text{otherwise}
        \end{cases}       
\end{equation}

\noindent
with $\boldsymbol{r_{a,obs}}$ the vector from the agent to the nearest border of each obstacle. Likewise, the goal generates a virtual attractive force from $f_{att, max}$ to $0$ with $d_{goal}$ the distance to start slowing down:

\begin{equation}
    \label{eq:4}
    \boldsymbol{f_{a,goal}} =
        \begin{cases}
        f_{att, max} & \text{if $\boldsymbol{\left \| r_{a,goal} \right \| } > d_{goal}$}\\
        f_{att, max} \cdot \frac{\boldsymbol{\left \| r_{a,goal} \right \| }}{d_{goal}} & \text{otherwise}
        \end{cases}       
\end{equation}

In our case, the goal will be each of the successive waypoints of a collision-free route calculated with a global planner. The joint action of all the virtual repulsive forces and the virtual attractive force gives a total force, $\boldsymbol{F_{E, C}}$, which represents the effect of the task's environment calculated at the frame $C$. Fig.~\ref{fig:force_based_model}~-~{\it A} shows a simplification of these forces calculation process for the robot alone.

\begin{equation}
	\label{eq:5}
	\boldsymbol{F_{E, C}} = w_{Rep} \cdot \left ( \sum_{obs=1}^{O-1} w_{obs} \cdot \boldsymbol{f_{C, obs}} \right ) + w_{Att} \cdot \boldsymbol{f_{C, goal}}
\end{equation}

Since the total virtual repulsive force's amplitude depends on the number of obstacles, it is necessary to normalize their addition using $w_{obs}$  ($\sum_{obs=1}^{O-1} w_{obs} = 1$) weighs in order to never exceed $f_{rep, max}$.

% Since the total virtual repulsive force's amplitude depends on the number of obstacles, it is necessary to normalize their addition using $w_{obs}$  ($\sum_{obs=1}^{O-1} w_{obs} = 1$) weighs to narrow down the result to $f_{rep, max}$ if the sum exceeds this value.

The idea behind the $w_{Rep}$ and $w_{Att}$ weights is that the maximum amplitude for the attractive, $f_{att, max}$, and repulsive, $f_{rep, max}$, forces is not necessary the same so they can be normalized to have a unique maximum value, $f_{max}$.

\begin{equation}
	\label{eq:6}
	\begin{aligned}
	    f_{rep,max} = f_{max} \cdot \frac{1}{w_{rep}}, ~ f_{att,max} = f_{max} \cdot \frac{1}{w_{att}} \\
	    f_{max} = w_{rep} \cdot f_{rep,max} = w_{att} \cdot f_{att,max}
	\end{aligned}
\end{equation}

Selecting the waypoints which generates each partial goal in such a way that there are no obstacles between the collaborative pair and the following goal, it can be inferred that $w_{Rep} < w_{rep}$ and $w_{Att} > w_{att}$ must be fulfilled to ensure that the maximum repulsive force is always smaller than the maximum attractive force to make the robot to move towards the next goal and do not oscillate. This ensures that the force component towards the goal is always positive in~(\ref{eq:5}) except if $\boldsymbol{\left \| r_{a,goal} \right \| } < d_{goal}$ when $\boldsymbol{F_{E, C}}$ could be $0$. This implies that the equilibrium point will be reached at a distance $d < d_{goal}$ closer or further to the goal depending on the number and on how close the obstacles are to it. 

Additionally, this environmental force can be normalized to impose that $f_{max}$ as maximum in order to make it comparable to the measured maximum human's exerted force if we select $f_{max} = f_{human, max}$.

% It can be inferred that $w_{Rep} > w_{rep}$ or $w_{Att} < w_{att}$ must be fulfilled to ensure that the maximum repulsive force can always become greater than the maximum attractive force to avoid possible collisions with obstacles. Additionally, this environmental force can be normalized to impose that $f_{max}$ as maximum in order to make it comparable to the measured maximum human's exerted force if we select $f_{max} = f_{human, max}$.

\begin{equation}
	\label{eq:7}
	\boldsymbol{F^{norm}_{E, C}} = \frac{\boldsymbol{F_{E, C}}}{w_{Rep} + w_{Att}} 
\end{equation}

\subsection{Human intention}

Whereas in the classic Perception-Action cycle the desired force to be applied would be calculated by directly adding the human's exerted force to $\boldsymbol{F^{norm}_{E, C}}$, we could use this last force as well as the virtual generated forces to understand the human's intention, which we postulate can be divided into two terms: implicit intention and explicit intention.

The first one is calculated as follows. The force exerted by the human on the other end of the object, $\boldsymbol{f_{human}}$, is detected at the robot's force sensor as the difference between the force exerted by each agent over the object:

\begin{equation}
	\label{eq:8}
	\boldsymbol{f_{sen}} = \boldsymbol{f_{human}} - \boldsymbol{f_{robot}} 
\end{equation}

If the robot moves without acceleration, the force due to this agent will be $0$. In fact, if both the human and the robot move at the same velocity and the human does not desire to accelerate or slow down, no force will appear at the sensor. In addition, the robot knows it own movement so it can discount its effect as well as the objects weight to calculate the force due to the human's action at the force sensor:

\begin{equation}
	\label{eq:9}
	\boldsymbol{f_{h,sen}} = \boldsymbol{R_{h, sen}} \cdot \boldsymbol{f_{human}}
\end{equation}

\noindent
being $\boldsymbol{R_{h, sen}}$ the transformation matrix which eliminates the robot's force and transforms the human's force from their grasping point to the robot's force sensor frame. Likewise, we can transform this force to frame $C$ using the corresponding transformation matrix, $\boldsymbol{R_{sen, C}}$:

\begin{equation}
	\label{eq:10}
	\boldsymbol{f_{h, C}} = \boldsymbol{R_{sen, C}} \cdot \boldsymbol{f_{h,sen}} = \boldsymbol{R_{sen, C}} \cdot \boldsymbol{R_{h, sen}} \cdot \boldsymbol{f_{human}}
\end{equation}

Finally, this force is compared with the attractive virtual force previously generated in (\ref{eq:4}) resulting in a coefficient proportional to the angle between the human's force in $C$ and the goal force, $\phi_{h,C;goal}$:

\begin{equation}
	\label{eq:11}
	i_{im} = k \cdot cos \left(\phi_{h,C;goal} \right)
\end{equation}

\noindent
with $k=1$ if it is detected that the human's force goes in favour of the task and with $k=0$ if it goes against the task. More details will be given in Section~\ref{sec:roles_and_strategies}.

As for the explicit intention, this is clearly related to the the informed intention previously mentioned in Section~\ref{sec:PIA_cycle}. This term depends on the subtask and on the means of communication used by the human. In general, we convert this explicit intention to a change in the environment. The general expression for this term is similar to~(\ref{eq:2}) but taking into the account that this term is valid for a limited period of time $t$.

\begin{equation}
	\label{eq:12}
	\boldsymbol{f_{ex,o}} \propto U_{a,o} \left ( \left \| \boldsymbol{r_{a,o}} \right \|,~t \right )
\end{equation}

We generate a explicit intention vector for each of the commands given by the human. For example, if the human indicates that they wish to avoid a particular path, this command will generate a virtual obstacle which generates an extra repulsive force. If they indicate that they wish to pass through a narrow passage, it will substitute the task's goal by a temporal subtask's goal at the other end of the passage generating an attractive force.

\subsection{Situation Awareness}

With all the virtual forces and coefficients calculated in the previous steps, we can understand the current situation introducing the situation awareness, to the best of our knowledge, for the first time in robotics. For that, this block considers if the implicit intention is relevant according to its coefficient and generates the human's contribution to the desired force:

\begin{equation}
	\label{eq:13}
	\boldsymbol{F_{H, C}} = \boldsymbol{f_{h, C}} + \boldsymbol{f_{im}} = \boldsymbol{f_{h, C}} + i_{im} \cdot \boldsymbol{f_{h, C}}
\end{equation}

\noindent
being $\boldsymbol{f_{im}}$ the implicit intention force when its coefficient is considered as relevant. 

We use this intention as a way to potentiate the human's force, as it can be seen in Fig.~\ref{fig:force_based_model}~-~{\it B}. The human's force is transformed to frame $C$ and the implicit intention coefficient is calculated. The total force potentiates the human's force avoiding the obstacle on the left while keep going towards the goal.

Likewise, the situation awareness converts the received commands into modifications of the attractive (generating subgoals) and repulsive (generating obstacles) forces taking into account their time validity and the previous understanding of the current situation. Fig.~\ref{fig:force_based_model}~-~{\it C} illustrates the case of a narrow passage through which the human has explicitly expressed their interest in going through.

Thus, the desired force to be applied to the object for each explicit intentions is:

% \begin{equation}
% 	\label{eq:12}
% 	\boldsymbol{F_{Des, obj_{l,s}}} = w_{E} \cdot \boldsymbol{F^{norm}_{E, obj_{l}}} + w_{H} \cdot \boldsymbol{F_{H, obj_{s}}}
% \end{equation}

% \noindent
% which can be reformulated in terms of the human's force:

% \begin{equation}
% 	\label{eq:13}
% 	\boldsymbol{F_{Des, obj_{l,s}}} = w_{E} \cdot \boldsymbol{F^{norm}_{E, obj_{l}}} + w_{h} \left( 1+i_{st,s} \right) \cdot \boldsymbol{f_{human}}
% \end{equation}

\begin{equation}
	\label{eq:14}
	\begin{split}
    \boldsymbol{F_{Des, C_{e}}} & = w_{E} \cdot \boldsymbol{F^{norm}_{E, C_{e}}} + w_{H} \cdot \boldsymbol{F_{H, C}} \\
    & = w_{E} \cdot \boldsymbol{F^{norm}_{E, C_{e}}} + w_{h} \left( 1+i_{im} \right) \cdot \boldsymbol{f_{h, C}}
    \end{split}
\end{equation}

\noindent
being $\boldsymbol{F^{norm}_{E, C_{e}}}$ the $e$ modified version of the task's environment force. Like in (\ref{eq:7}), the desired force can also be normalized using $w_{E}+w_{H}$ instead in order to delimit it to $f_{max} = f_{human, max}$. 

\begin{equation}
	\label{eq:15}
	\boldsymbol{F^{norm}_{Des, C_{e}}} = \frac{\boldsymbol{F_{Des, C_{e}}}}{w_{E} + w_{H}} 
\end{equation}

This process is repeated with each explicit intention $e \in I_e$ being $I_e$ the set of considered intentions. Finally, each desired force can be used to generate a future projection of how the object will move and its impact on the development of the task.

\section{Robot's Platform Controller}\label{sec:controller}

\begin{figure*}[t]
    \centering
    \includegraphics[width=0.98\textwidth]{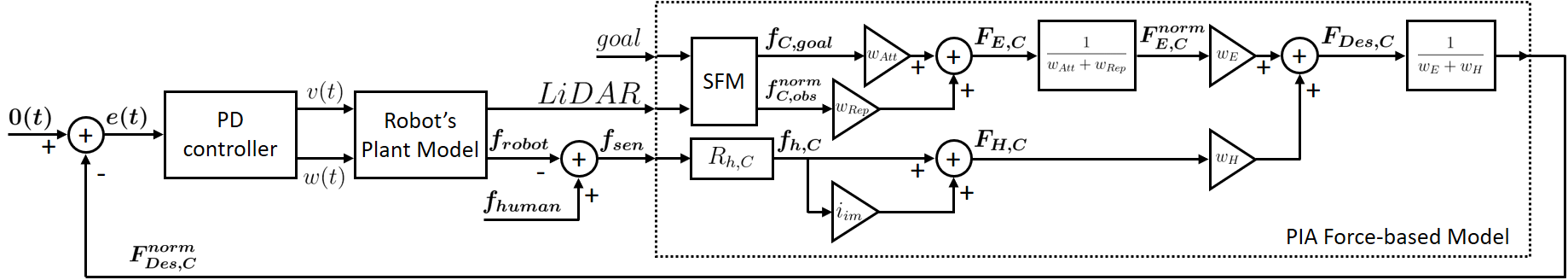}
    \caption{{\bf Control structure used to generate the platform's velocity commands.} A PD controller is used to generate the linear and angular velocity from the desired force result of the joint action of the environment and human forces. These commands are sent to the robot's internal controller making the robot to do whatever is necessary to fulfill the task taking into account the human's intention.}
    \label{fig:PD_controller}
\end{figure*}

Considering that in the previous approach the robot moves on a plane, the calculated resultant force, $\boldsymbol{F^{norm}_{E, C}}$, will have two components, $x$ and $y$, which can be used to calculate the linear and angular velocity that the robot must have. Fig.~\ref{fig:PD_controller} shows a diagram of the control cycle used to achieve this.

The logic of operation is as follows. While the task is running, there will be a goal and potentially obstacles in the way. This added to the choice of weights in (\ref{eq:6}), guarantees that $\boldsymbol{F_{E,C}}$ is not null. At the same time, being a collaborative transport task, the human will have to make an effort (higher or smaller depending on the role they assume but some effort after all) making $\boldsymbol{F_{H,C}}$ not null either unless the human perfectly follows the robot's movements. This implies that $\boldsymbol{F_{Des, C}}$ will only be $0$ when the task is finished. Therefore, we take $0$ as the set-point and $\boldsymbol{F_{Des, C}}$ as a perturbation. The difference is sent to a task controller (in our case, a PD controller) which generates the linear and angular velocity commands necessary for the robot to perform the task by adapting in the process to the human's wishes. This also ensures that, if the human exerts the exact opposite force to cancel the $\boldsymbol{F_{E,C}}$ component by opposing the task and making $\boldsymbol{F_{Des, C}} = 0$, the robot stops.

Note that we split the platform control into two blocks: our task controller and the robot's internal wheel controller which would be inside the plant model. This may seem like a weakness of our model, but it is actually a strength as it allows us to decouple our controller from the robot's own controller. This allows us to abstract from the specific robot we are using and use our model in any other robot.

\section{Collaborative Task Roles and Strategies}\label{sec:roles_and_strategies}

A Markov decision process can take care of the generated future projections to decide about the next actions to be performed. Even a simple finite state machine can do it if we only consider the most relevant intentions and, therefore, work with only one projection.

The first step is to determine the role or the sharing policy used by each agent. For that, we start with the classical roles of {\it master}, {\it slave} and {\it collaborative} exposed in works like~\cite{Mortl2012} but we also seek inspiration in works like~\cite{Jarrasse2012} or~\cite{Losey2018} considering two new roles.

We define the {\it collaborative} role as the one exerted by team members who consider each other as equals and who contribute with their experience and skills (not necessarily equal among them) to achieve a better performance of the task. Similarly, we define the {\it master} role as the one performed by the agent who imposes their will and their vision of how the task should be performed correctly, and we assign the {\it slave} role to the agent who accepts the will of a master and fulfills their plans. Note that in all three cases, the agents intend to accomplish the task. The same is not true for the {\it neutral} and {\it adversarial} roles we propose. We assign a {\it neutral} role to the agent who neither acts in favour of the task nor against its correct performance. Finally, we consider an agent as an {\it adversary} if their intention goes against accomplishing the task at hand.

Examples of these roles can be found in Fig.~\ref{fig:force_based_model}. In situation B, the human exerts such a force that the robot is able to interpret that the human's desire is to go towards the goal but making an effort to avoid the obstacle on the left. The robot accepts this intention, assigns a collaborative role to both and potentiate the human's intention. This can be mathematically interpreted as increasing the $w_{H}$ pre-eminence over $w_{E}$ by $i_{im}$. Fig.~\ref{fig:force_based_model}~-~{\it C} shows a situation where previously the human has indicated their intention to cross a narrow passage. The robot accepted their will and made a modification not over the weights but over $\boldsymbol{F^{norm}_{E, C_{e}}}$ itself and, since to cross the narrow passage both must collaborate (stand one behind the other), a collaborative role is assigned to both for the duration of this interaction. Finally, Fig.~\ref{fig:force_based_model}~-~{\it D} shows an example of detecting that the human's intention goes against the task. The robot rejects this will, assigns an adversarial role to the human and stops taking the human's intention into account in its calculations. In this way, we potentiate the $w_{E}$ weight.

%Examples of these roles can be found in Fig.~\ref{fig:force_based_model}. In situation B, the human exerts such a force that the robot is able to interpret that the human's desire is to leave the object on what appears to be an obstacle next to the original goal. The robot accepts this intention and assigns a role of master to the human and slave to itself. This can be mathematically interpreted as increasing the $w_{H}$ pre-eminence over $w_{E}$ by $i_{st,s}$. Fig.~\ref{fig:force_based_model}~-~{\it C} shows a situation where previously the human has indicated their intention to cross a narrow passage. The robot accepted their will and made a modification not over the weights but over $\boldsymbol{F^{norm}_{E, obj_{e}}}$ itself and, since to cross the narrow passage both must collaborate (stand one behind the other), a collaborative role is assigned to both for the duration of this interaction. Finally, Fig.~\ref{fig:force_based_model}~-~{\it D} shows an example of detecting that the human's intention goes against the task. The robot rejects this will, assigns an adversarial role to the human and stops taking the human's intention into account in its calculations potentiating this way the $w_{E}$ weight.

These are just some of the possible strategies that the robot may decide to follow to handle each situation. It could put both weights to zero in the adversarial case or opt for a progressive increase of each weight in the master and slave cases respectively. It could also decide to try other communication means like natural language to try to understand the human's intention. In any case, the decision will be dependent on the robot's knowledge of the task.

% These are just some of the possible actions that the robot may decide to perform in each situation since there are other possible strategies to handle each situation. The robot can, for example, be more restrictive in accepting the human's intention and be their slave in situation B or allow the human to lead the way in situation C since they are the one who wants to go through that passage. Similarly, it can also choose to stop completely until the human behaves in a different way in situation D or even try to communicate with it using voice commands and natural language processing.

% Additionally, the level of uncertainty associated with each future projection can be used to make the robot decide to opt for more cautious actions until it can reduce that uncertainty. In any case, the decision will be dependent on the robot's knowledge of the task.

\section{Experiments}\label{sec:experiments}

A preliminary round of experiments should be done to prove all of the previous considerations. For that, three volunteers (age: $\mu=25.67$, $\sigma=2.89$; most common ongoing or finished studies: M.Sc.) performed up to $15$ experiments ($5$ each one) in which the robot and the human perform a collaborative transportation task through different scenarios with multiples obstacles. 

\subsection{Experiments Setup}

The first two experiments are for the human to learn the robot's capabilities: in the first, the robot constantly assumes the role of teacher (ignoring the force exerted by the human) so that the human learns the robot's navigation capabilities. In the second, the robot assumes the role of slave throughout the experiment (it overrides the goal force and avoids colliding with obstacles) so that the human discovers how to operate the robot as well as its response speed. In the third and fourth, along the shortest route to the goal there is a hidden forbidden path sign that only the human can recognize. This makes the human have to force the robot to follow another route. In the third experiment, they will have only their own strength to do so, while in the fourth experiment they will be given the possibility to explicitly express their intention. Finally, the fifth experiment will involve a narrow passage on the shortest route that is impossible to cross unless one of the agents stands behind the other. As in the fourth experiment, the human will also be given the possibility to express their intention to cross the narrow passage. Fig.~\ref{fig:human-robot_pair_experiment} shows a case of the latter experiment.

To allow the human to explicitly indicate their intention, we have designed a handle with 5 buttons, one for each finger, allowing the first three to (1) take control of the robot (robot as slave), (2) indicate that you want to cross a narrow passage and (3) indicate that the current route is not allowed (the last two buttons have no assigned functionality).

As for the robot used, it is a TIAGo++\footnote{\url{https://pal-robotics.com/robots/tiago/}} manufactured by PAL Robotics. The weights used are $w_{Att} = 1.01$, $w_{Rep} = 0.99$, $w_E = w_h = 1.0$. The controller gains are $k_P = -2.0$ and $k_D = -0.35$. The stage used is an indoor area of $7.8$ x $5.7~m$ with OptiTrack on the ceiling to allow localization of both agents. After each experiment, the human fills out a questionnaire to thus obtain objective and subjective data.

\subsection{Validation}

\begin{figure}[t]
    \centering
    \includegraphics[width=0.98\textwidth]{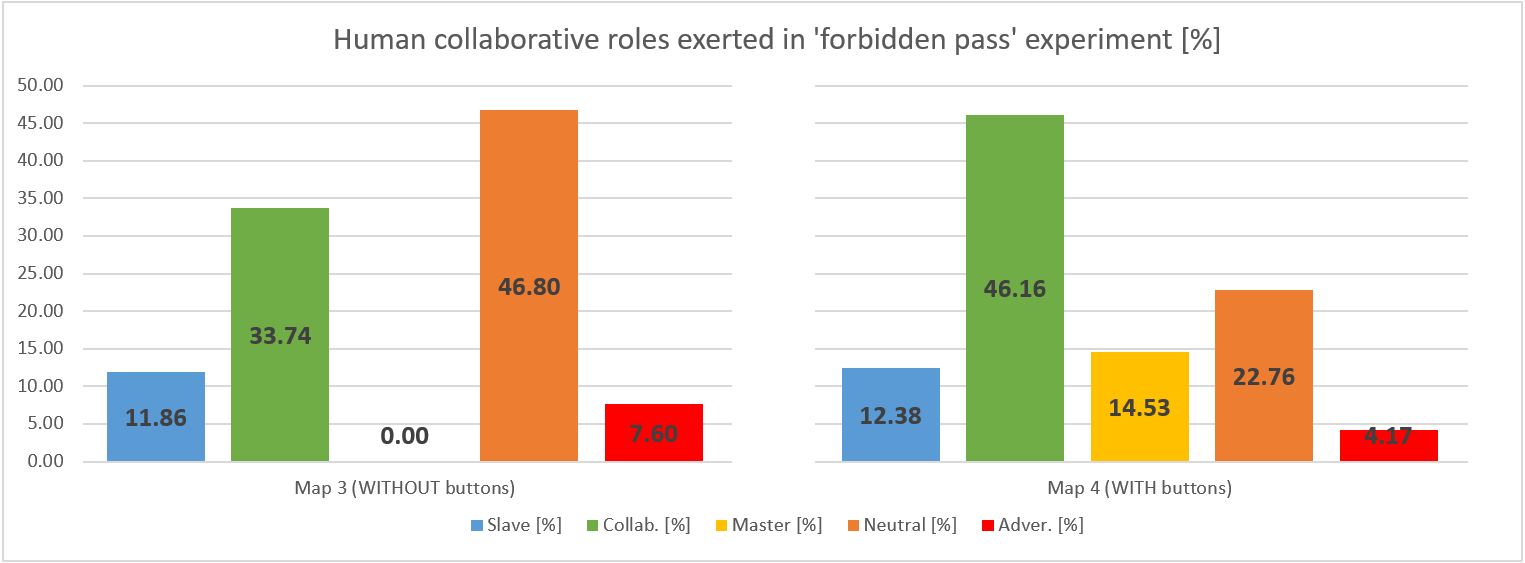}
    \caption{{\bf Comparison of the roles exerted by the voluntaries in the third and forth experiment.} Percentage of each role detected with our system. Buttons in the handle are disabled in map 3 and enabled in map 4.}
    \label{fig:Human_roles_map_3_vs_4}
\end{figure}

\begin{figure}[t]
    \centering
    \includegraphics[width=0.49\textwidth]{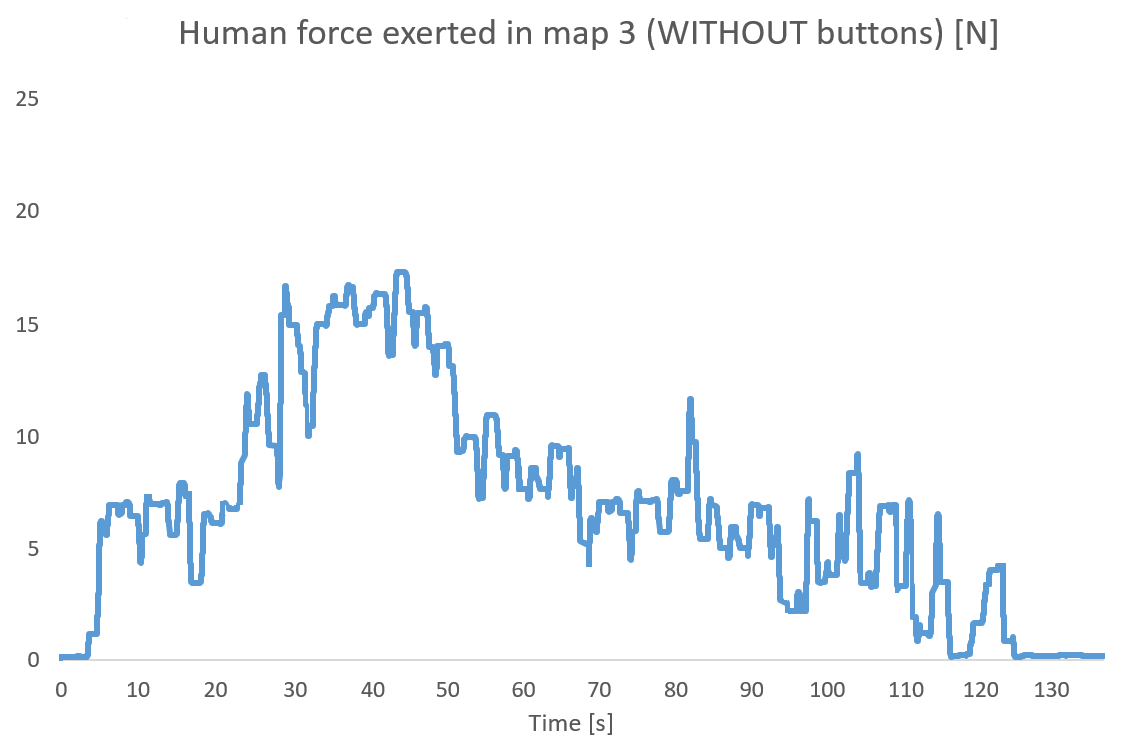}
    \includegraphics[width=0.49\textwidth]{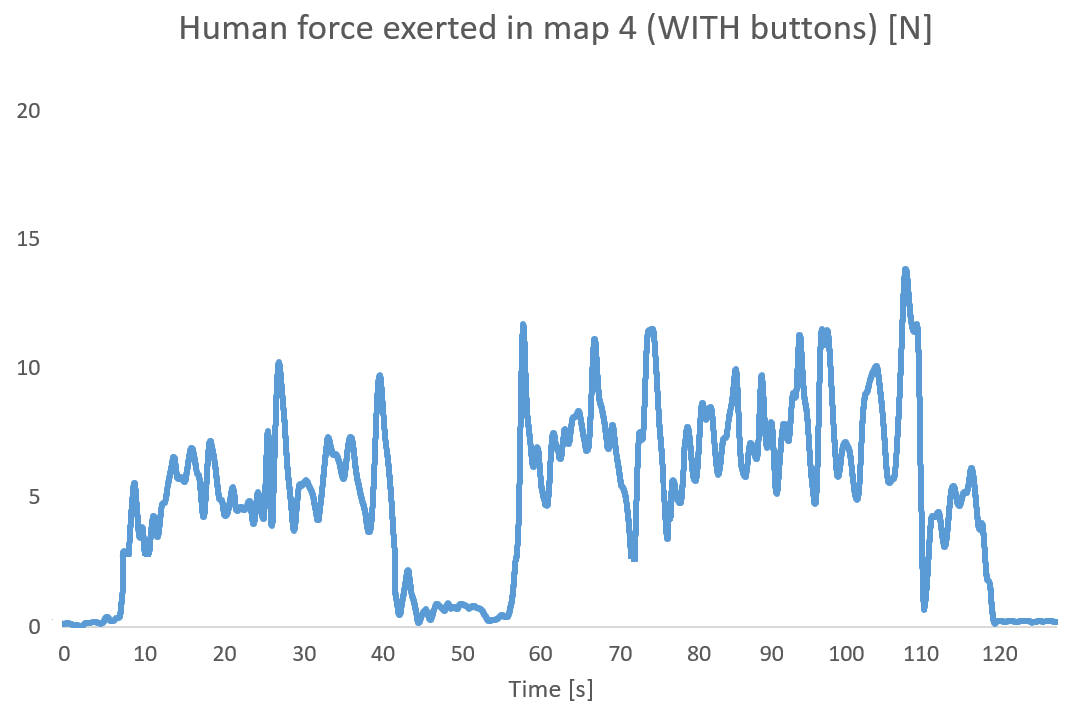}
    \caption{{\bf Evolution of the force exerted by the voluntaries in the third and forth experiment.} Extra force needed in map 3 once the forbidden path sign is seen to make the robot to go backwards until it replans using other route. No extra force needed in map 4.}
    \vspace{0mm}
\label{fig:Human_force_map_3_vs_4}
\end{figure}

The first two experiments are to give the user a certain minimum skill, so for the sake of brevity they will not be analyzed here. If we analyze the third and fourth experiments which serve as a direct comparison between having and not having a way of explicitly expressing the human's intention, in Fig.~\ref{fig:Human_roles_map_3_vs_4} we can observe the difference which occurs when assigning a role to the human. In the third experiment, the human is not able to indicate to the robot that it should not follow a specific route and is forced to exert a force which the robot considers to be that of an adversarial agent or simply uncatalogable (neutral). In the fourth experiment, on the other hand, the human does have a way of explicitly indicating that the robot should not follow that route (and another way of taking control of the robot directly). This allows the robot to understand their intentions and assign them a collaborative role most of the time (or master if the human chooses to impose themselves). The evolution of the force exerted can be seen in Fig.~\ref{fig:Human_force_map_3_vs_4}. Note the difference in scale in both graphs.

\begin{figure}[t]
    \centering
    \includegraphics[width=0.35\textwidth]{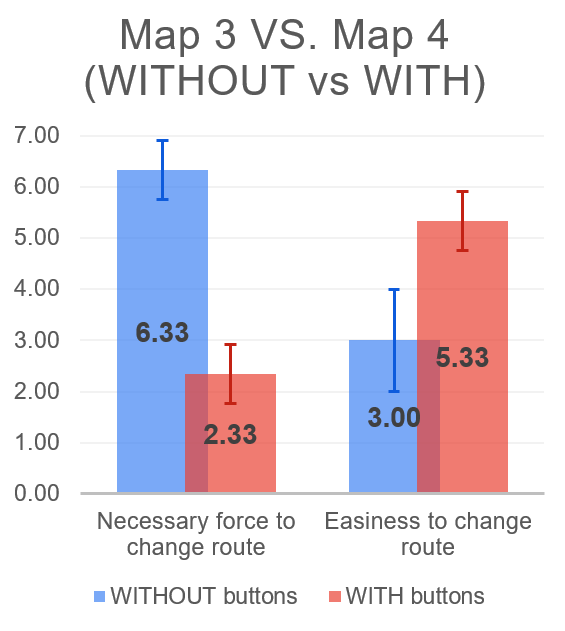}
    \includegraphics[width=0.61\textwidth]{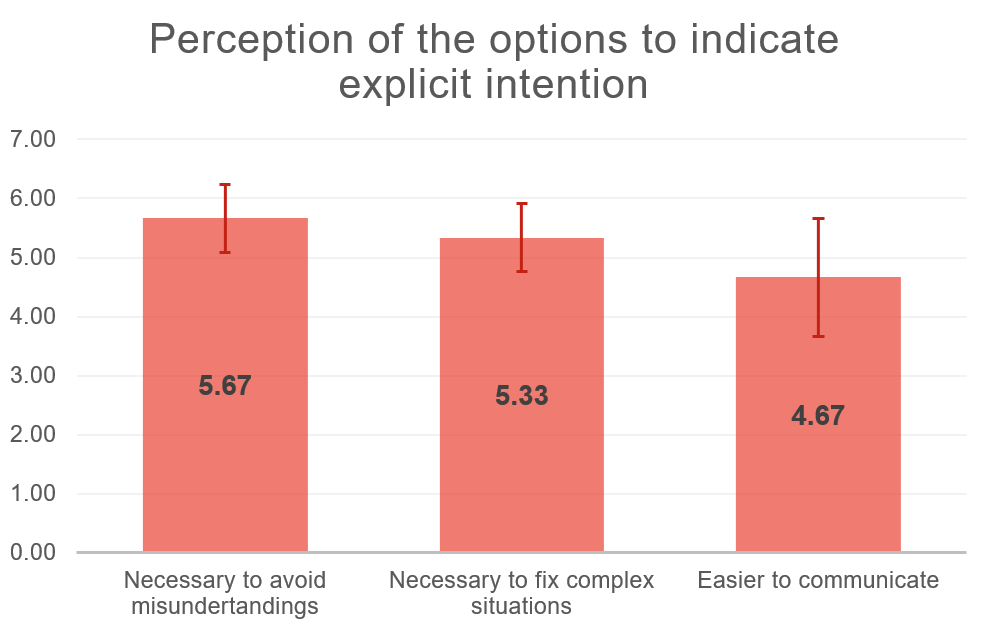}
    \caption{{\bf User study.} Comparison of the difficulty to impose their intention in maps 3 and 4 and general evaluation of the utility of the explicit intention.}
    \vspace{0mm}
\label{fig:User_study}
\end{figure}

This difference is also perceived by the human as reflected in the questionnaires completed at the end of each experiment (Fig.~\ref{fig:User_study}), indicating that it is much easier for them to indicate what they want in Experiment 4. The last questionnaire completed by the volunteers confirms that they understand that it is necessary to explicitly indicate their intention in order to avoid misunderstandings and to solve situations which would be difficult to solve otherwise. In turn, they also find it easier to communicate with the robot.

\section{Conclusions and Future Work}\label{sec:conclusions}

We have reviewed the perception-action cycle adding the human's intention to it using for that the concept of situational awareness. We have also modeled this new complete cycle using a force-based model as a way to take into consideration the human activity in collaborative human-robot tasks. For that we have combine virtual and physical forces in a mathematically easy to understand and implement model utilisable in any mobile robot. Finally, we have carried out a preliminary round of experiments to check that the human understand the necessity of telling their explicit intention and that they feel comfortable working with the robot and understood by its internal model. 

Further analysis of the post-experiment questionnaires can give insight about the human's preferences. Additionally, more complex architectures to generate future projections like neural networks and extra information inputs like the human's gaze can be explored to obtain more adaptive collaborative plans.

\section*{Acknowledgments}

The authors want to express their gratitude to Ferr\'{a}n Cort\'{e}s and Patrick Grosch for their technical support and to the volunteers who made this work possible.

\bibliographystyle{IEEEtran}
\balance
\bibliography{IEEEabrv,./hri.bib}

\end{document}